\documentclass{article}

 \PassOptionsToPackage{numbers, compress}{natbib}
\usepackage[final]{nips_2017}

\usepackage[utf8]{inputenc} % allow utf-8 input
\usepackage[T1]{fontenc}    % use 8-bit T1 fonts
\usepackage{hyperref}       % hyperlinks
\usepackage{url}            % simple URL typesetting
\usepackage{booktabs}       % professional-quality tables
\usepackage{amsfonts}       % blackboard math symbols
\usepackage{amsmath}
\usepackage{nicefrac}       % compact symbols for 1/2, etc.
\usepackage{microtype}      % microtypography
\usepackage{subfiles}
\usepackage{xcolor}
\usepackage{multirow}
\usepackage{enumerate}
\usepackage{subfiles}
\usepackage{multirow}
\usepackage{graphicx}
\usepackage{subfiles}

\newcommand{\graphlongcapitalized}{Program-Derived Semantics Graph}
 
\newcommand{\graphlong}{program-derived semantics graph}
\newcommand{\langlong}{program-derived semantics language}

\newcommand{\graph}{PSG}
\newcommand{\lang}{PSL}

\title{Software Language Comprehension using a Program-Derived Semantics Graph}

\begin{document}
\bibliographystyle{abbrv}
\maketitle
\vspace{-7mm}
\begin{abstract}
\vspace{-3mm}
Traditional code transformation structures, such as abstract syntax trees (ASTs), conteXtual flow graphs (XFGs), and more generally, compiler intermediate representations (IRs), may have limitations in extracting higher-order semantics from code. While work has already begun on higher-order semantics lifting (e.g., Aroma's simplified parse tree (SPT), verified lifting's lambda calculi, and Halide's intentional domain specific language (DSL)), research in this area is still immature. To continue to advance this research, we present the \emph{\graphlong} (\graph), a new graphical structure to capture semantics of code. The \graph\ is designed to provide a single structure for capturing program semantics at multiple levels of abstraction. The PSG may be in a class of emerging structural representations that cannot be built from a traditional set of predefined rules and instead must be \emph{learned}. In this paper, we describe the \graph\ and its fundamental structural differences compared to state-of-the-art structures. Although our exploration into the \graph\ is in its infancy, our early results and architectural analysis indicate it is a promising new research direction to automatically extract program semantics.
% \blfootnote{Code available at \url{https://github.com/tensorflow/tensor2tensor}}

%TODO(noam): update results for new models.

%llion@: FAIR's paper seems to concentrate solely on the convolutional aspect of their model and have the attention as an after thought almost, this gives us a good opportunity to differentiate ourselves from their paper.

%We are simpler in a number of ways and should have the simplicity as a big selling point:
%\begin{itemize}
%\item No convolutions
%\item No need for such careful initializations and %normalization.
%\item Simpler non-lineararities, they use the gated linear %units.
%\item Less layers?
%\end{itemize}
%One thing we do more is that we have self attention.
%Another selling point is the increased interpretability as %shown with the visualizations. Which comes from the %simplicity and use of only attentions.
\end{abstract}

\vspace{-5mm}

\section{Introduction}
\label{sect:introduction}

\vspace{-4mm}

\emph{Machine programming} (MP), defined as any system that automates some portion of software, envisions a future where machine learning (ML) can (nearly) automate the entire software development lifecycle~\cite{gottschlich:2018:mapl}. A core open challenge in MP is the ability to automatically extract user intention from code~\cite{ye:2020:misim}. Exacerbating this problem, new programming languages (PLs) continue to be developed with varying levels of semantic abstraction (e.g., Halide, Python, and SYCL)~\cite{keryell:2015:iwocl, ragan-kelley:2013:pldi, rossum:2009:python3}. Such semantic variability may handicap traditional single dimensional hierarchical structures, such as ASTs, which can generally only represent code at a semantic level for which the syntax exists. Further, these structural limitations might create potential inconsistency and incompatibility in semantic representations from one PL to the next. In this paper, we aim to address this problem with a new structure called the \emph{\graphlong} (\graph). The PSG's principle purpose is to capture program semantics. However, different from prior structures of which we are aware (e.g., AST, XFG, SPT, etc.) it achieves this in a novel way by introducing a hierarchical structure that varies semantic granularity. The \graph\ is also graphical in nature which we leverage to identify relationships that might be challenging (or impossible) to represent with tighter constraints such as a tree, which does not allow for certain characteristics like cycles. To capture the richness and nuances of abstract semantic concepts that may be difficult to precisely define by rules, we envision the PSG to pioneer research towards developing \emph{learned} precise programming structural representations. The PSG's representation has numerous applications in software engineering including code translation between PLs, bug detection and root-cause mitigation, code question-answering, and code optimization. 

\begin{figure}[htp]
    \centering
    \includegraphics[width=8.5cm, height=9.5cm]{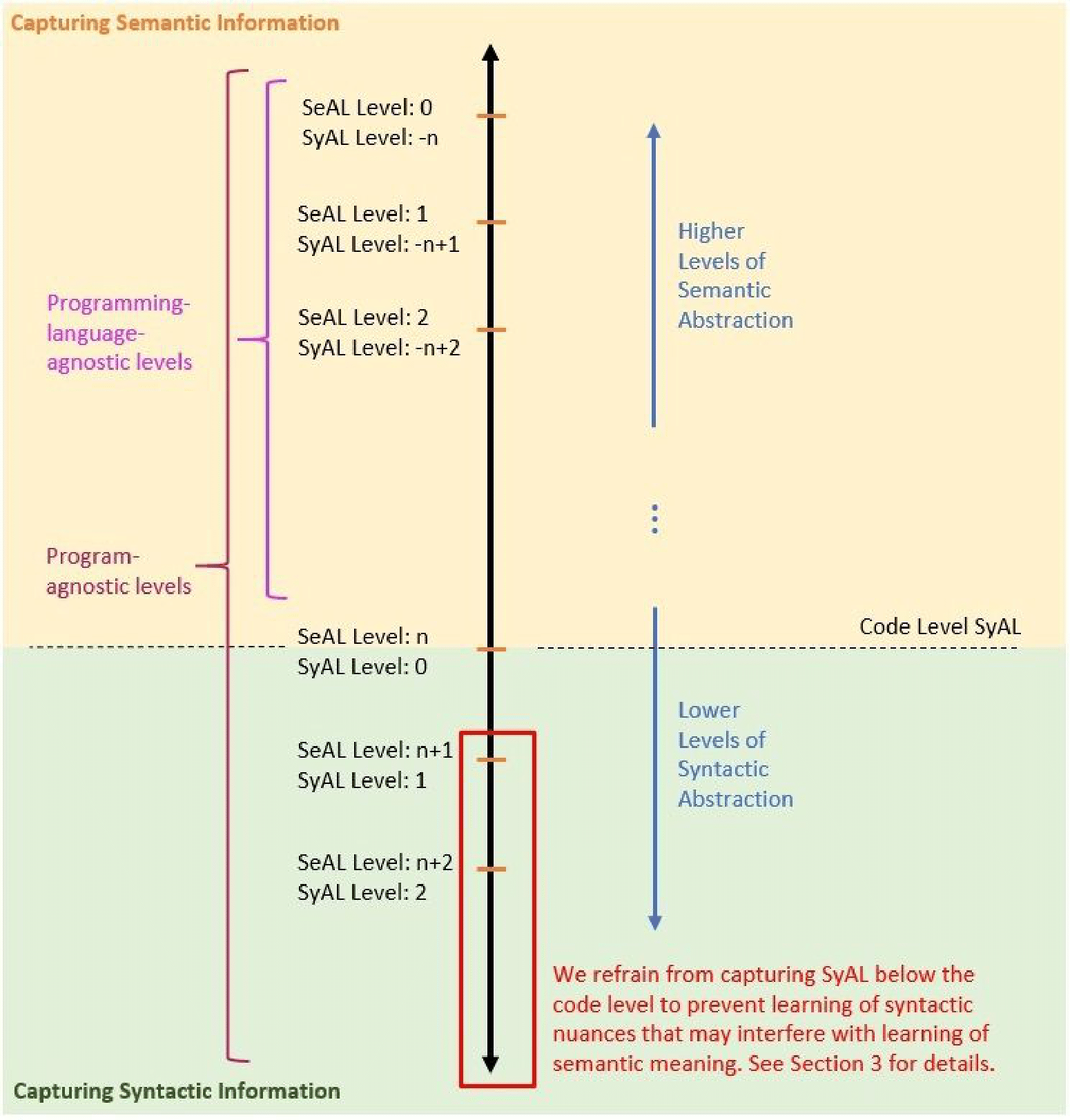}
    \caption{\textmd{\graph\ Abstraction Level Spectrum for Semantic Abstraction Levels (SeAL) and Syntactic Abstraction Levels (SyAL), distinguished by color-coding.}}
    \label{fig:asg_spectrum}
\end{figure}

\vspace{-3mm}

\section{Related Work}
\label{section:related_work}

\vspace{-4mm}

In this section, we discuss recent efforts to extract code semantics and how they compare to our \graph.

\vspace{-3mm}

\paragraph{Verified Lifting.}
\emph{Verified lifting} (VL) is a technique that analyzes code from a source PL, lifts its semantics to a higher level representation, then lowers it into a target PL~\cite{kamil:2016:pldi}. However, a core challenge with VL may be in its semantic abstraction. Though VL uses a single-level DSL to store semantics, there may be cases where a hierarchical abstraction system is required when moving code between PLs due to their differing abstraction levels (AL). Further, mapping DSL semantic abstractions to hierarchical ALs may improve VL's ability to extract code semantics.

\vspace{-2.8mm}

\paragraph{Neural Code Comprehension.}
The goal of the Neural Code Comprehension (NCC) system is to extract code semantics using a fusion of programmatic structural representations in addition to ML-based modeling~\cite{nun:2018:neurips}. NCC introduces a novel structure called the conteXtual flow graph (XFG) to extract semantic meaning through identified data dependencies. The XFG, however, is limited to languages supported by low level virtual machine intermediate representations (LLVM-IR) for which it is lowered to. Moreover, as the XFG is principally grounded to lower-level syntactic representations, its structure may have challenges in extracting the underlying semantic meaning of code.  

\vspace{-2.8mm}

\paragraph{Aroma.}
Aroma is a novel code recommendation system~\cite{luan:2019:oopsla} that emphasizes learning semantics, rather than syntax, through a code's structure using an SPT. We structurally analyze the \graph\ and Aroma's SPT, which indicates that the PSG may learn more semantics.

\vspace{-3mm}

\section{Program-Derived Semantics Graph (\graph) \& Language (\lang)}

\vspace{-3mm}

\label{section:extracting_semantic_meaning}
\subsection{\graphlongcapitalized}
\label{subsection:psl_and_psg}

\vspace{-3mm}

The \graph\ is a multi-tiered representation of program semantics derived from a program's source code. The \graph\ is PL-agnostic and is a graphical representation of our hierarchical abstract semantic concept language, the \lang. We have designed the \lang\ in hopes of it being both adaptable and extensible. The Appendix summarizes our motivation for representing the \graph\ as a graph data structure.

\begin{figure}[htp]
    \centering
    \includegraphics[width=10cm, height=7cm]{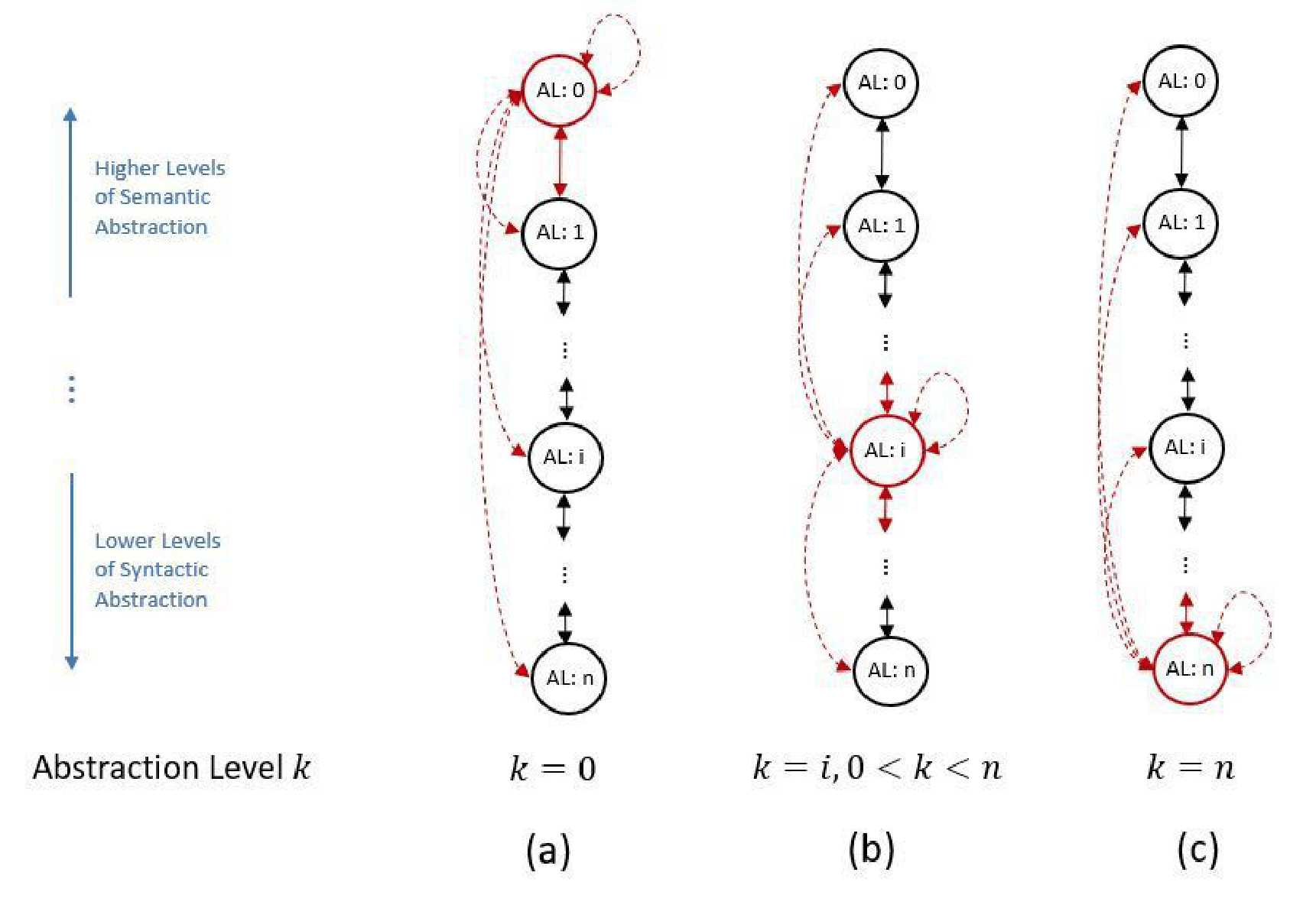}
    \caption{\textmd{\graph\ AL Dependencies with potential dependencies (dotted red arrows), and minimum dependencies (solid red arrows); (a) shows dependencies for the highest AL, $k = 0$. (b) shows dependencies for intermediate ALs, $0 < k < n$. (c) shows dependencies for the lowest AL, $k = n$.}}
    \label{fig:ASG_dependencies}
\end{figure}

\noindent
\emph{The \graph\ Structure.}
The \graph\ incorporates both semantic and syntactic information through hierarchical ALs.~\footnote{Appendix Figure \textcolor{red}{\ref{fig:base_ASG}} provides a detailed graphical representation of our base \lang, known as base \graph. We adopt the term \emph{base \graph} because it uses a first-order approximation \lang, base \lang. For future work, we aim to provide a more comprehensive construction of the base \lang, by mining previously unseen code structures.}
As illustrated in Figure \textcolor{red}{\ref{fig:asg_spectrum}}, each \graph\ level provides a varying degree of granularity. Higher levels of abstraction capture more abstract and general semantic information, while lower levels of abstraction encode more syntactic and precise information. Providing the correct level of representation of code syntax for semantic analysis is difficult. It continues to be an open challenge in compilers~\cite{mlir:2019:google}, which often perform too much syntactic analysis that they can obfuscate a system's ability to extract semantic meaning. In the Appendix, we describe an example to motivate the need for finding a balance between syntactic and semantic analysis. Our current embodiment of the \graph\ only incorporates one layer of concrete syntactic abstraction in \emph{SyAL Level: 0}, which is PL-specific. Like Aroma before us, we made this decision to avoid capturing too many nuances of PL syntax, which may interfere with the goal of the PSG to capture program meaning, not implementation details. At the same time, our aim with \emph{SyAL Level: 0} is to help address the problem of learning syntactic information to extract out semantic meaning through categorical classification.

%Semantic overlap for Figures (a) and (b) are indicated in purple. (c) SPT construction for Implementation 1. (d) SPT construction for Implementation 2. Semantic overlap for Figures (c) and (d) are indicated in blue.}}

\vspace{-3mm}

\noindent
\paragraph{\graph\ is Adaptable and Extensible.} As we are in the early stages of exploring the \graph, it may be unwise to provide a static structure for the \graph. Moreover, as PLs evolve and new PLs are created, our aim is for the \graph\ to automatically and seamlessly handle such dynamic changes. To address these challenges, we have designed the \graph\ to be \emph{adaptable} and \emph{extensible}. In this way, we believe the PSG may also likely be the first concrete solution to automatically enhance software even as PLs evolve. The \graph\ is adaptable in that it is PL-agnostic, and further, as PLs evolve, the \graph\ is designed to evolve with them with its support of the syntactic layer.  The \graph\ is also extensible since its layers of abstraction are not fixed. This is to accommodate the potential need for adding more layers in order to capture an evolving range of algorithms. The Appendix describes how large-scale software data may be a key component to comprehensively and automatically maitain the PSL and PSG.
\vspace{-3mm}

\subsection{Structural Analysis}
\label{subsection:scalability}

\vspace{-3.5mm}

Due to the \graph's graphical nature and base \lang's design, we believe the \graph\ may be a scalable structure to represent code bodies that have been historically limited~\cite{luan:2019:oopsla}, for the following reasons: First, the \graph\ contains a single AL for capturing syntactic information, with all other levels capturing semantic concept information. Moreover, our base \lang\ does not include details such as variable or function names. Eliding such information has the byproduct of reducing an instantiation of the \graph's overall size. Second, as the levels of the \graph\ are raised, the size of ALs shrink.  This is because lower levels of abstraction expand upon higher order AL categories by enumerating all objects belonging to those categories. As such, we would expect for the lowest AL, which encodes for syntactic abstraction at the code-level, to be the largest level of abstraction in the \graph. The Appendix provides a conceptual illustration. While prior work that has extracted semantics solely from syntactic information has usually been restricted to inputs of a dozen lines of code or less, we believe it may be possible using the \graph's scalable approach to increase the input sizes.

As a demonstration of the \graph's structure, we show an example of how it captures semantics from two code snippets that are semantically equivalent but syntactically different. We consider two possible implementations of exponentiation (i.e., $x^y$) in C++. One performs the operation recursively, and the other performs it iteratively. The reader is referred to the Appendix for implementation details. We generate the \graph\ for both implementations, shown in the Appendix, which we refer to as \graph-recursive and \graph-iterative respectively. We also generate Aroma's SPT of each implementation, shown in the Appendix, which we refer to as SPT-recursive and SPT-iterative respectively.

We perform a rudimentary analysis of the resulting \graph s and SPTs,  measuring their corresponding node overlap percentages and differences using the method described below.~\footnote{Though not intended to be comprehensive, our analysis is one approach for computing semantic similarity.} For this analysis, we refer to the multiset of nodes in a given structure (in this case, either a \graph\ or SPT) as $N$. For the purposes of this analysis, there are two multisets to consider, which we refer to as $N_1$ and $N_2$. The below analysis is performed twice: once for the resulting \graph s and then again for the resulting SPTs.

\vspace{-2mm}
\begin{enumerate}

\item Calculate multiset intersections: $I_1 = N_1 \cap N_2$ and $I_2 = N_2 \cap N_1$.~\footnote{$N$ represents a multiset. As such, the intersection of $N_1 \cap  N_2$ is not necessarily equivalent to $N_2 \cap N_1$.}

\vspace{-1mm}
\item Calculate percentages of intersection: $P_1 = (|I_1| \div |N_1|)$ and $P_2 = (|I_2| \div |N_2|)$.

\vspace{-1mm}
\item Calculate percentages distance (degree of difference): $\eta = |P_1 - P_2|$

\vspace{-1mm}
\item Calculate approximate lower bound:
  $L = |min(P_1, P_2) - \eta|$

\vspace{-1mm}
\item Calculate range, $R$, and average, $A$, of structural similarity:\\ 
  $R = [L, min(P_1, P_2)]$, $A = (L + min(P_1, P_2)) \div 2$.

\end{enumerate}

\vspace{-2mm}
Shown above, we devise two novel calculations intended for code similarity. Those are the calculation of (3) the percentages distance, $\eta$, and (4) an approximate lower bound, $L$, that are explained below.

In our analysis of code similarity systems, we note that one code snippet, $S_1$, may have a large percentage intersection with another code snippet, $S_2$. Yet, $S_2$'s intersection with $S_1$ may be small. This irregularity difference is notable because it implicitly argues two opposing views: \emph{(i)} the code snippets are similar and \emph{(ii)} the code snippets are dissimilar. Logically, both views cannot simultaneously hold. To capture these differences, we introduce $\eta$ which grows the greater the difference between the two percentages of intersection. We then introduce $L$, an approximate lower bound, as a penalty for such a difference. The intuition behind $L$ is that similarity overlaps that are relatively small percentage differences (i.e., a small $\eta$) should be penalized less than similarity overlap percentages with relatively large percentage differences (i.e., a large $\eta$). We find that fusing these two calculations presents one possible analysis of both the potential similarity between two code snippets, and of potential limitations of a structural representation used to extract semantic meaning.

Next, we compute the semantic structural similarity from the above procedure between SPT-recursive and SPT-iterative, and between \graph-recursive and \graph-iterative. The computation details are described in the Appendix. For the SPT, we find that $R = [63.70\%, 64.71\%]$ and $A = 64.21\%$. For the PSG, we find that $R = [69.91\%, 70.37\%]$ and $A = 70.14\%$. From this analysis, the \graph\ defines these code representations to be on average $5.93\%$ more structurally similar compared to the SPT. Although these results are anecdotal, we believe they provide early intuition on how the \graph\ may be used and how it generally compares, structurally, to Aroma's SPT. Construction of the PSG and our measurement of semantics does not study a program by its execution or use program synthesis techniques such as code input and output analysis. Rather, our technique constructs and analyzes a PSG without requiring code compilation, a practical assumption in many cases for code development. 

\vspace{-2mm}

\section{Conclusion and Future Work}
\label{section:conclusion}

\vspace{-4mm}

This position paper provides a conceptual framework for reasoning about code semantics through a \graphlong\ (PSG). In it, we presented the concept and intuition of both the \graph\ and the \langlong\ (\lang). The \graph\ is a graphical representation of our hierarchical abstract semantic concept language, the \lang, which we have designed in the hopes of it being both adaptable and extensible. We described their fundamental structure and intuition, and illustrated both the \graph\ and \lang\ against Aroma's state-of-the-art simplified parse tree.

For future work, we plan to analyze the \graph\ and other state-of-the-art systems, like Aroma, against larger code corpora. We also leave the problem of automatically generating the \graph\ from code as part of future work that is currently work-in-progress. While we have early ideas for tackling this problem, the system design is outside the scope of this paper.

\newpage
\bibliographystyle{plain}
%\bibliography{deeplearn}

% \bibliography{main}

\begin{thebibliography}{10}

% \bibitem{layernorm2016}
% Jimmy~Lei Ba, Jamie~Ryan Kiros, and Geoffrey~E Hinton.
% \newblock Layer normalization.
% \newblock {\em arXiv preprint arXiv:1607.06450}, 2016.

\bibitem{mlir:2019:google}
Google.
\newblock Multi-Level \uppercase{IR} \uppercase{C}ompiler \uppercase{F}ramework. 
\newblock {\em https://www.tensorflow.org/mlir}, 2019.

\bibitem{nun:2018:neurips}
Tal~Ben-Nun, Alice~Shoshana Jakobovits, and Torsten Hoefler.
\newblock Neural Code Comprehension: A Learnable Representation of Code Semantics.
\newblock {\em Advances in Neural Information Processing Systems 32},
\newblock NeurIPS 2018, pages 3585–3597. Curran Associates, Inc., 2018.

\bibitem{berghammer:2016:card}
Rudolf~Berghammer, Nikita~Danilenko, Peter~H{\"{o}}fner, and Insa~Stucke.
\newblock Cardinality of relations with applications.
\newblock {\em Discret. Math}, 339(12):3089–3115, 2016.

\bibitem{cosentino:2017:ieee}
Valerio~{Cosentino}, Javier~L. {Cánovas Izquierdo}, and Jordi {Cabot}.
\newblock A Systematic Mapping Study of Software Development With GitHub.
\newblock {\em IEEE Access}, 5:7173–7192, 2017.

\bibitem{gottschlich:2018:mapl}
Justin~Gottschlich, Armando~Solar-Lezama, Nesime~Tatbul, Michael~Carbin, Martin~Rinard, Regina~Barzilay, Saman~Amarasinghe, Joshua~B. Tenenbaum, and Tim~Mattson.
\newblock The Three Pillars of Machine Programming.
\newblock In {\em Proceedings
of the 2nd ACM SIGPLAN International Workshop on Machine Learning and Programming
Languages}, MAPL 2018, pages 69–80, New York, NY, USA, 2018. ACM.

\bibitem{kamil:2016:pldi}
Shoaib~Kamil, Alvin~Cheung, Shachar~Itzhaky, and Armando~Solar-Lezama.
\newblock Verified Lifting of Stencil Computations.
\newblock In {\em Proceedings of the 37th ACM SIGPLAN Conference on Programming Language Design and
Implementation},
\newblock PLDI ’16, pages 711–726, New York, NY, USA, 2016. ACM.

\bibitem{keryell:2015:iwocl}
Ronan~Keryell, Ruyman~Reyes, and Lee~Howes.
\newblock  Khronos SYCL for OpenCL: a tutorial.
\newblock In {\em Proceedings of
the 3rd International Workshop on OpenCL},
\newblock IWOCL’15, Palo Alto, CA, USA, 2015. Association
for Computing Machinery.

\bibitem{kipf:2017:iclr}
Thomas~N. Kipf and Max~Welling.
\newblock  Semi-supervised Classification with Graph Convolutional Networks.
\newblock In {\em Proceedings of the International Conference on Learning Representations},
\newblock ICLR, 2017.

\bibitem{luan:2019:oopsla}
Sifei~Luan, Di~Yang, Celeste Barnaby, Koushik~Sen, and Satish~Chandra.
\newblock Aroma: Code Recommendation via Structural Code Search.
\newblock {\em Proc. ACM Program. Lang.},
\newblock  3(OOPSLA):152:1–152:28, Oct 2019.

\bibitem{ragan-kelley:2013:pldi}
Jonathan~Ragan{-}Kelley, Connelly~Barnes, Andrew~Adams, Sylvain~Paris, Fr{\'{e}}do~Durand, and Saman~P. Amarasinghe.
\newblock Halide: A Language and Compiler for Optimizing Parallelism, Locality, and Recomputation in Image Processing Pipelines.
\newblock In {\em ACM SIGPLAN Conference on Programming Language Design and
Implementation, PLDI ’13, Seattle, WA, USA, June 16-19, 2013},
\newblock pages 519–530, 2013.

\bibitem{schlichtkrull:2019:iclr}
Michael~Schlichtkrull, Thomas~N. Kipf, Peter~Bloem, Rianne~van den Berg, Ivan~Titov, and Max~Welling
\newblock Modeling Relational Data with Graph Convolutional Networks.
\newblock In {\em Proceedings of the International Conference on Learning Representations},
\newblock ICLR, 2019.

\bibitem{rossum:2009:python3}
Guido~ Van Rossum and Fred~L. Drake
\newblock {\em Python 3 Reference Manual}.
\newblock CreateSpace, Scotts Valley, CA,
2009.

\bibitem{ye:2020:misim}
Fangke~Ye, Shengtian~Zhou, Anand~Venkat, Ryan~Marcus, Nesime~Tatbul, Jesmin~Jahan Tithi, Paul~Petersen, Timothy~Mattson, Tim~Kraska, Pradeep~Dubey, Vivek~Sarkar, and Justin~Gottschlich
\newblock MISIM: An End-to-End Neural Code Similarity System, 2020.

\end{thebibliography}

\clearpage
\appendix

\section*{Appendix}
\label{section:appendix}    

\section{Motivation for a Graphical Representation}

We were motivated to represent the \graph\ as a graph data structure for some of the following reasons:

\begin{enumerate}

\item Graphs can effectively encode structural information, or preserve syntactic meaning, through parent-child-sibling node hierarchy. While both graphs and trees can preserve hierarchical structural information, graphs are more general. This generality may be useful when working on an open research question, like code similarity, where added flexibility may result in a broader exploration of solutions. 

\item Graphs can be effective representations for graph neural networks (GNNs) used to learn latent features or semantic information. Relational graph convolutional networks (R-GCNs)~\cite{schlichtkrull:2019:iclr} are a class of GNNs that apply graph convolutions on highly multi-relational graphs, like the \graph, to learn graph structure and semantic meaning. Models using GNN-based approaches have achieved promising results in the domain of representation learning~\cite{kipf:2017:iclr}.

\item The semantics of certain software abstraction levels may be more easily represented using a graph. One concrete example of this is illustrated in Neural Code Comprehension~\cite{nun:2018:neurips}. As shown in their contextual flow graph, dependencies of data and control flow may take on a graph structure, where two nodes can be connected by more than one edge. The authors show that a tree structure would be insufficient for capturing such (potentially) cyclic dependencies.

\end{enumerate}

\section{Example: The delicate balance between syntax and semantic analysis}
Consider a program that seeks to manipulate strings. In many languages, one could store such information in many ways. For example, the following are some ways one could implement a string in C++: \texttt{std::string}, \texttt{char[]}, \texttt{wchar\_t[]}, \texttt{std::vector<char>},\\ \texttt{std::array<wchar\_t, size>}, to name a few. If a semantic extraction system focuses too deeply on such implementation nuance, it may learn that two programs, which are identical in terms of semantics, are dissimilar due to divergent implementation details. On the other hand, ignoring all such details may eliminate information that is critical in interpreting the semantics of the program. For example, capturing the semantic details, through syntactic interpretation that a single variable is being used to manipulate information, rather than a collection of variables, could be critical in understanding cardinality constraints of a particular problem~\cite{berghammer:2016:card}.

It is for these reasons that the \graph\ captures some syntactic information to tailor semantic learning to the PL-specific functionalities as determined by the syntax. The last abstraction level of the \graph\ is PL-specific. Our aim with this layer is to help to address the problem of learning syntactic information to extract out semantic meaning through categorical classification.

\section{Structural Analysis of the PSG}
\begin{figure}[htp]
    \centering
    \includegraphics[width=9cm, height=7cm]{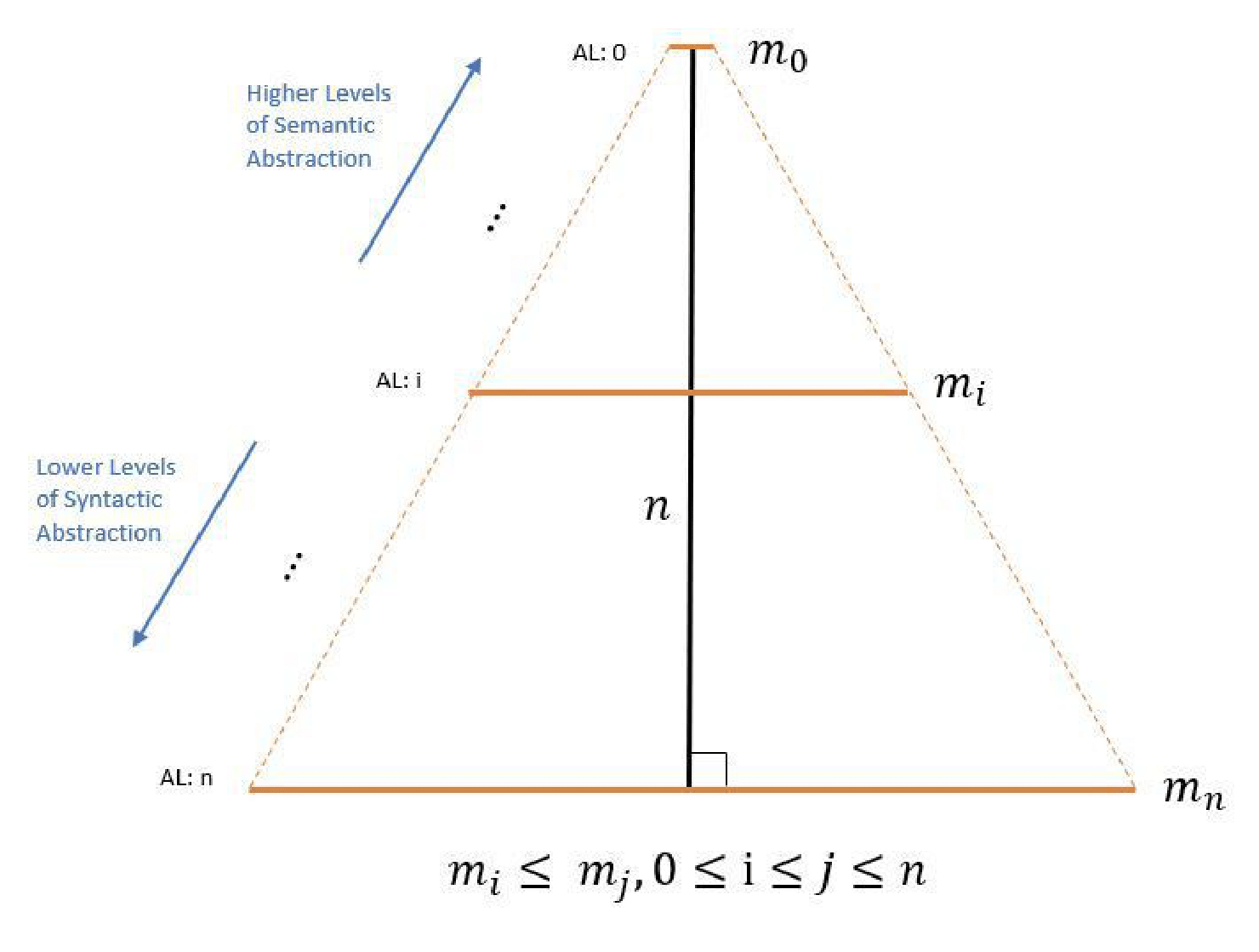}
    \caption{\textmd{Relationship between \graph\ Abstraction Level (AL) size and count. There are $n$ ALs where the size of each AL $i$ is $m_{i}$ which indicates the number of semantic concepts AL $i$ captures.}}
    \label{fig:asg_size_count}
\end{figure}

As shown in Figure \textcolor{red}{\ref{fig:asg_size_count}}, when the levels of the PSG are raised, the size of abstraction level $i$, denoted as $m_{i}$, shrinks. This is because lower levels of abstraction expand upon higher order abstraction level categories by enumerating all objects belonging to those categories. As such, we would expect for the lowest abstraction level, which encodes for syntactic abstraction at the code-level, to be the largest level of abstraction in the \graph.

\begin{figure}[t!]
    \centering
    \includegraphics[width=8cm, height=8cm]{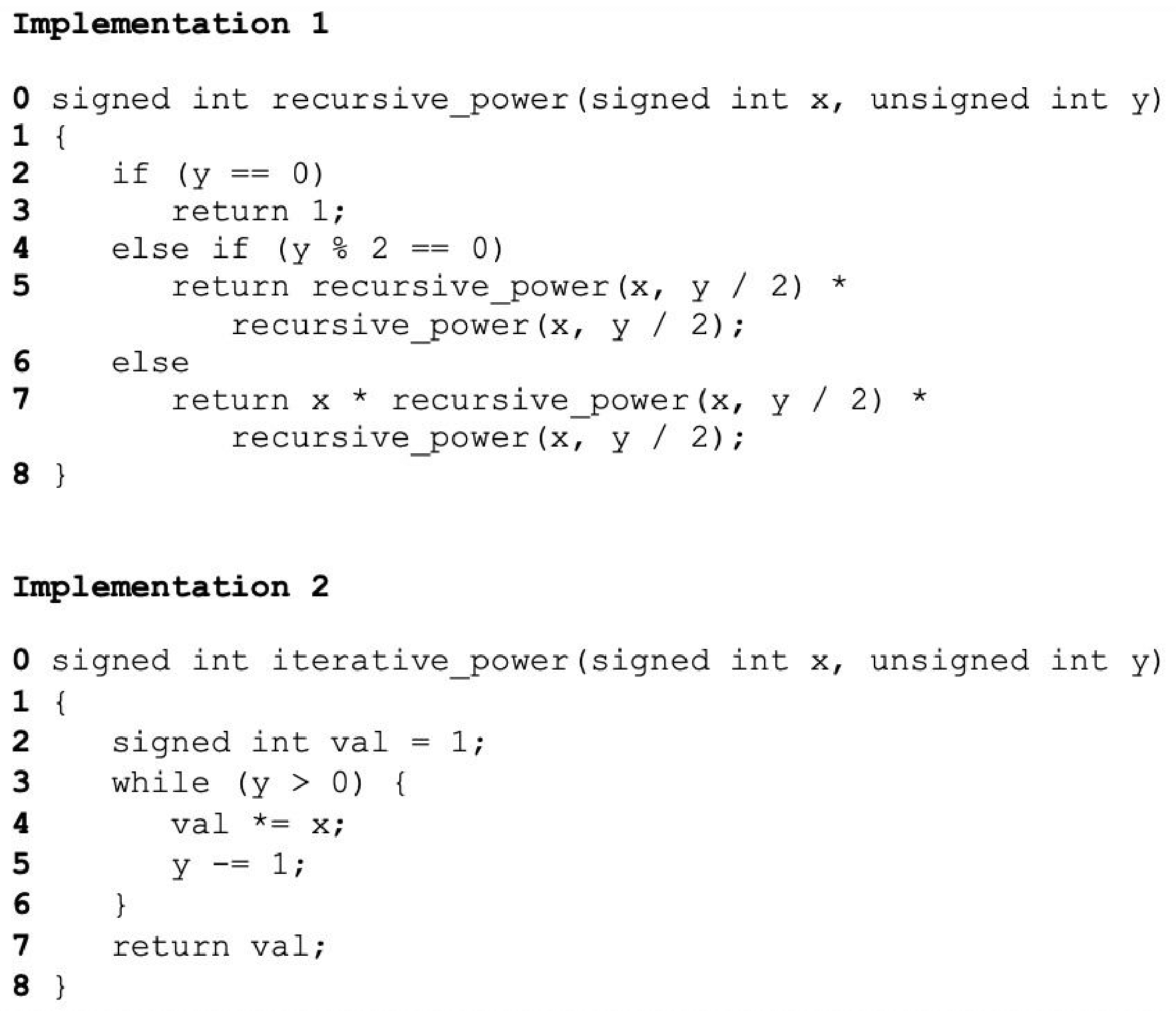}
    \caption{\textmd{Two functions computing exponentiation (i.e., $x^{y}, x \in \mathbb{R}$ and $y \geq 0$) in C++. Implementation 1 is recursive. Implementation 2 is iterative. The functions are semantically equivalent but syntactically inequivalent.}}
    \label{fig:x^y_program}
\end{figure}

\begin{figure*}[htp]
    \centering
    \includegraphics[width=16cm, height=9cm]{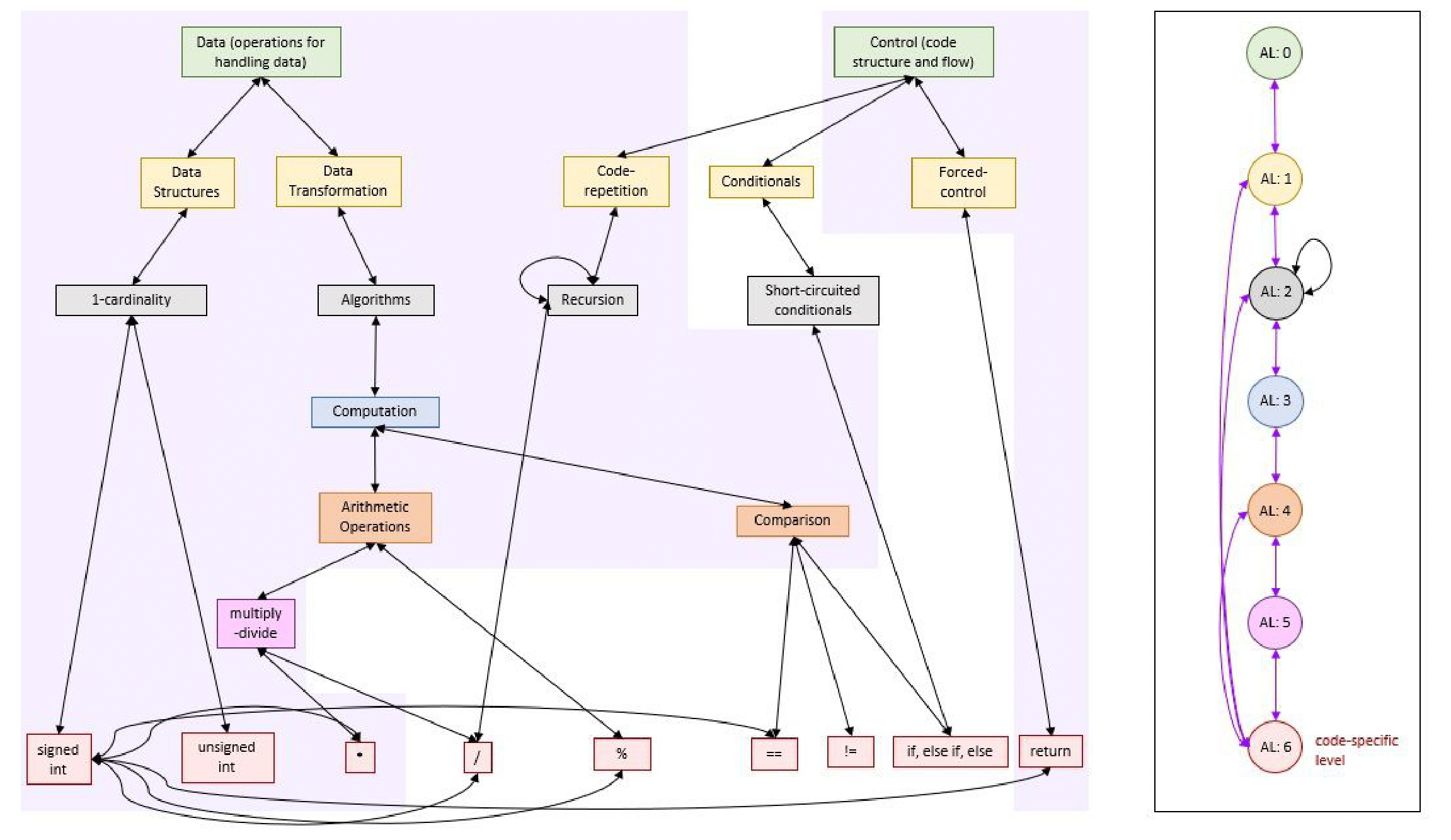}
    \caption{\textmd{\graph\ of Recursive Power Function. The shaded region denotes overlap in the nodes of the \graph\ for the iterative power function shown in Figure~\textcolor{red}{\ref{fig:asg_iterative}}. These total 17 of the 24 total nodes, a $70.83\%$ overlap.}}
    \label{fig:asg_recursive}
\end{figure*}

\begin{figure*}[htp]
    \centering
    \includegraphics[width=16cm, height=9cm]{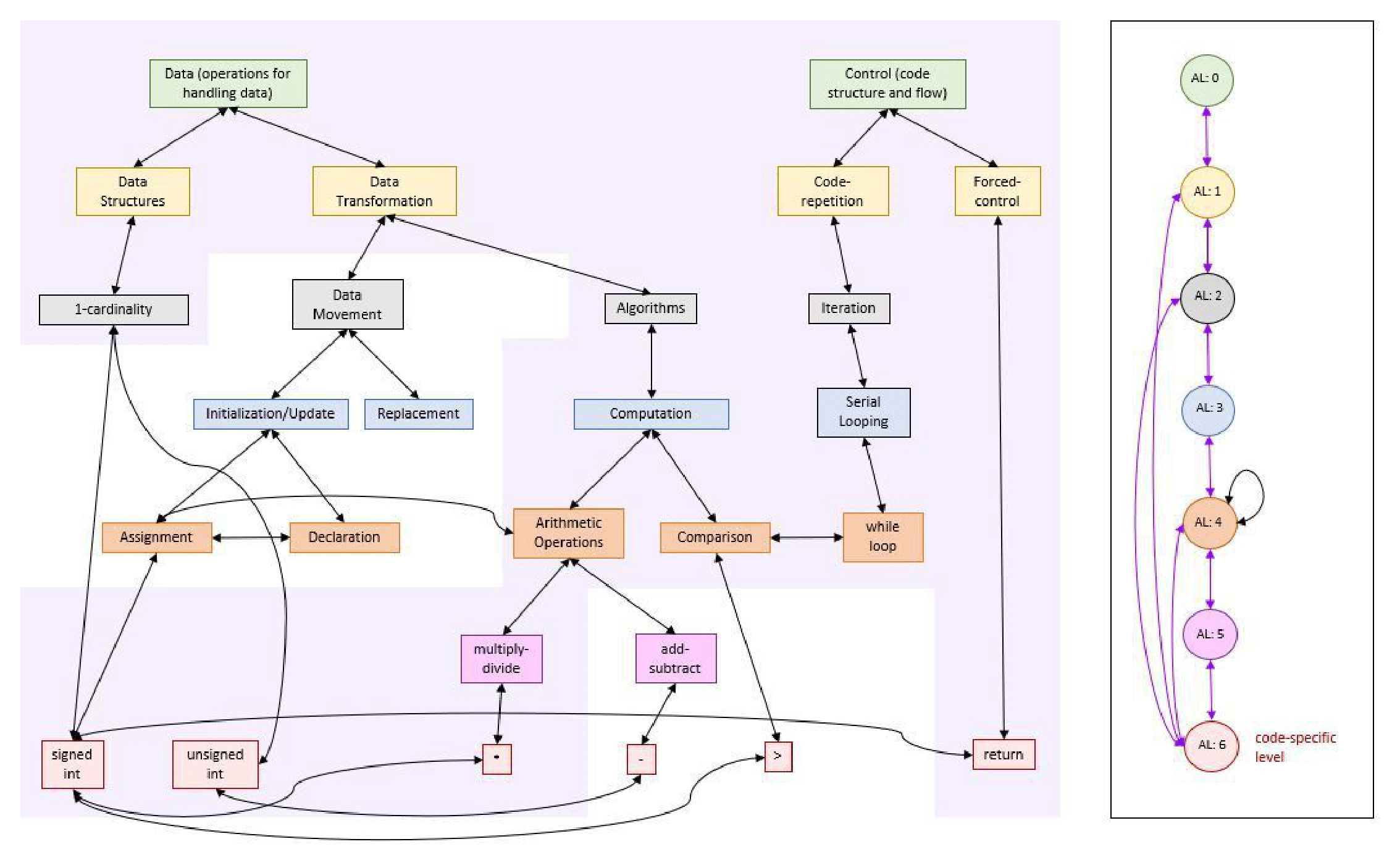}
    \caption{\textmd{\graph\ of Iterative Power Function. The shaded region denotes overlap in the nodes of the \graph\ for the recursive power function shown in Figure~\textcolor{red}{\ref{fig:asg_recursive}}. These total 19 of the 27 total nodes, a $70.37\%$ overlap.}}
    \label{fig:asg_iterative}
\end{figure*}

\begin{figure*}[htp]
    \centering
    \includegraphics[width=14cm, height=10cm]{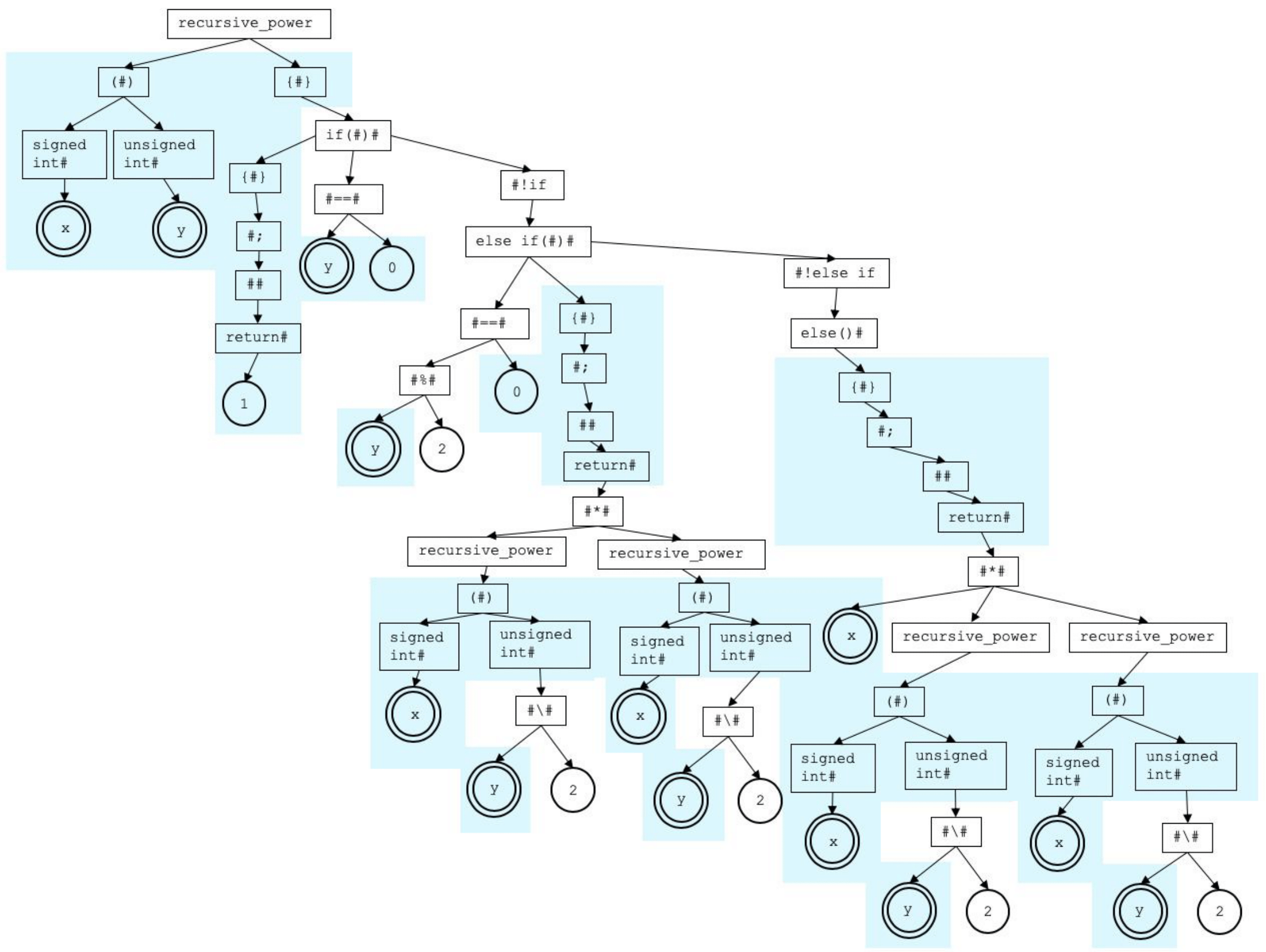}
    \caption{\textmd{SPT of Recursive Power Function. The shaded region denotes overlap in the nodes of the SPT for the iterative power function shown in Figure~\textcolor{red}{\ref{fig:spt_iterative}}. These total 44 of the 68 nodes, a $64.71\%$ overlap.}}
    \label{fig:spt_recursive}
\end{figure*}

\begin{figure*}[htp]
    \centering
    \includegraphics[width=14cm, height=9cm]{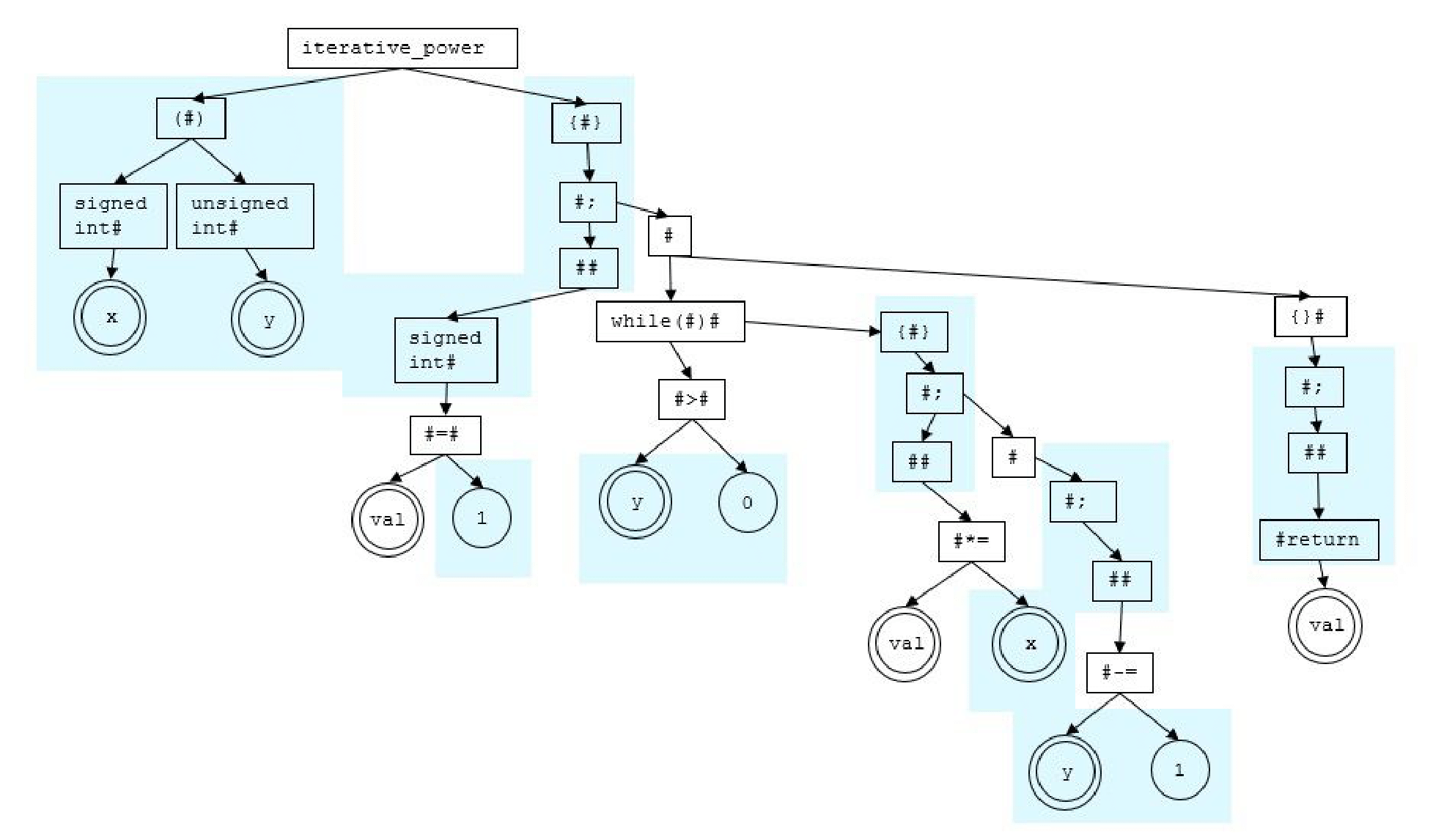}
    \caption{\textmd{SPT of Iterative Power Function. The shaded region denotes overlap in the nodes of the SPT for the recursive power function shown in Figure~\textcolor{red}{\ref{fig:spt_recursive}}. These total 23 of the 35 nodes, a $65.71\%$ overlap.}}
    \label{fig:spt_iterative}
\end{figure*}

As a demonstration of the \graph's structure, we show an example of how it captures semantics from two code snippets that are semantically equivalent but syntactically different. Further, using this example, we compare the \graph\ to Aroma's SPT. We consider two possible implementations of exponentiation (i.e., $x^y$) in C++, shown in Figure~\textcolor{red}{\ref{fig:x^y_program}}. One performs the operation recursively, and the other performs it iteratively. We generate the \graph\ of both implementations shown in Figures \textcolor{red}{\ref{fig:asg_recursive}} and \textcolor{red}{\ref{fig:asg_iterative}}. We refer to these representations as \graph-recursive and \graph-iterative. We also generate Aroma's SPT of each implementation in Figures \textcolor{red}{\ref{fig:spt_recursive}} and \textcolor{red}{\ref{fig:spt_iterative}}. We refer to these representations as SPT-recursive and SPT-iterative. 



\clearpage
\section{Structural Analysis Computation: SPT and PSG}
\noindent
For SPT: 

\begin{enumerate}
\setlength\itemsep{1.5em}
\item
$|N_{1}| = 68$, $|N_{2}| = 35$, $|I_1| = 44$, and $|I_2| = 23$

\item 
$P_1 = (|I_1| \div |N_1|)$  $= \frac{44}{68}$, $P_2 = (|I_2| \div |N_2|)$  $= \frac{23}{35}$

\item
$\eta = |P_{1} - P_{2}|$ $= \frac{6}{595}$

\item
$L = |min(P_1, P_2) - \eta|$ $= \frac{379}{595}$

\item
$R = [L, min(P_1, P_2)]$ \boxed{$= [63.70\%, 64.71\%]$}

%$= [\frac{379}{595}, \frac{44}{68}]$ \boxed{$= [63.70\%, 64.71\%]$}

 $A = (L + min(P_1, P_2)) \div 2$ \boxed{$= 64.21\%$}
\\
\end{enumerate}

\noindent
For \graph: 

\begin{enumerate}
\setlength\itemsep{1.5em}
\item 
$|N_{1}| = 24$, $|N_{2}| = 27$, $|I_1| = 17$, and $|I_2| = 19$
\item 
$P_1 = (|I_1| \div |N_1|)$ $= \frac{17}{24}$,
$P_2 = (|I_2| \div |N_2|)$ $= \frac{19}{27}$
\item
$\eta = |P_{1} - P_{2}|$ $= \frac{1}{216}$
\item
$L = |min(P_1, P_2) - \eta|$ $= \frac{151}{216}$
\item
$R = [L, min(P_1, P_2)]$ \boxed{$= [69.91\%, 70.37\%]$}
%$= [\frac{151}{216}, \frac{19}{27}]$ \boxed{$= [69.91\%, 70.37\%]$}

$A = (L + min(P_1, P_2)) \div 2$ \boxed{$= 70.14\%$}
\\
\end{enumerate}

\subsection{Learning the \lang\ With Data}
\label{section:data}

\vspace{-3mm}

Data may be a key component to enable a comprehensive and automatically maintained \lang. While base \graph\ is a working graphical representation of the first-order approximation \lang, it is non-exhaustive of all semantic concepts and their dependencies. To mitigate this weakness, we believe it may be possible to augment the base \lang\ through a continuously refined learning system, which will aim to learn new semantic concepts and dependencies of PLs from data patterns (i.e., anomalies) it has not previously observed. With the emergence of publicly large available code repositories (e.g., as of this writing, GitHub has around 200 million repositories~\cite{cosentino:2017:ieee}), and the growing magnitude of the web itself, we believe an automated and synthesized comprehensive \lang\ may be within our technological reach.

\clearpage
\section{Base PSG Abstraction Levels}

\begin{figure}[htp]
    \centerline{\includegraphics[scale=0.21]{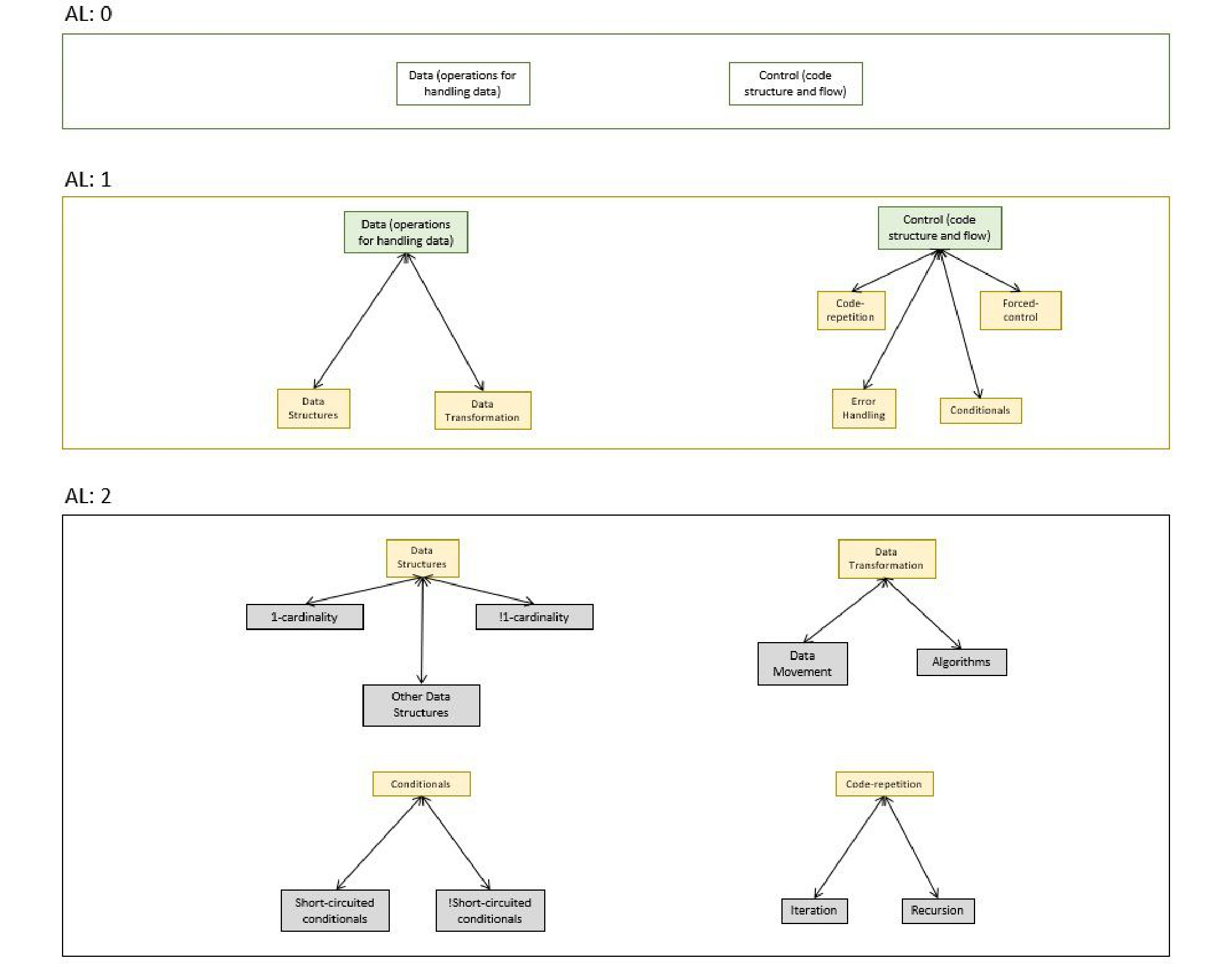}}
    % \centering
    % \includegraphics[width=16cm]{fig/fig_5_1.jpg}
\end{figure}

\label{section:appendix}
\begin{figure}[htp]
    \centerline{\includegraphics[scale=0.21]{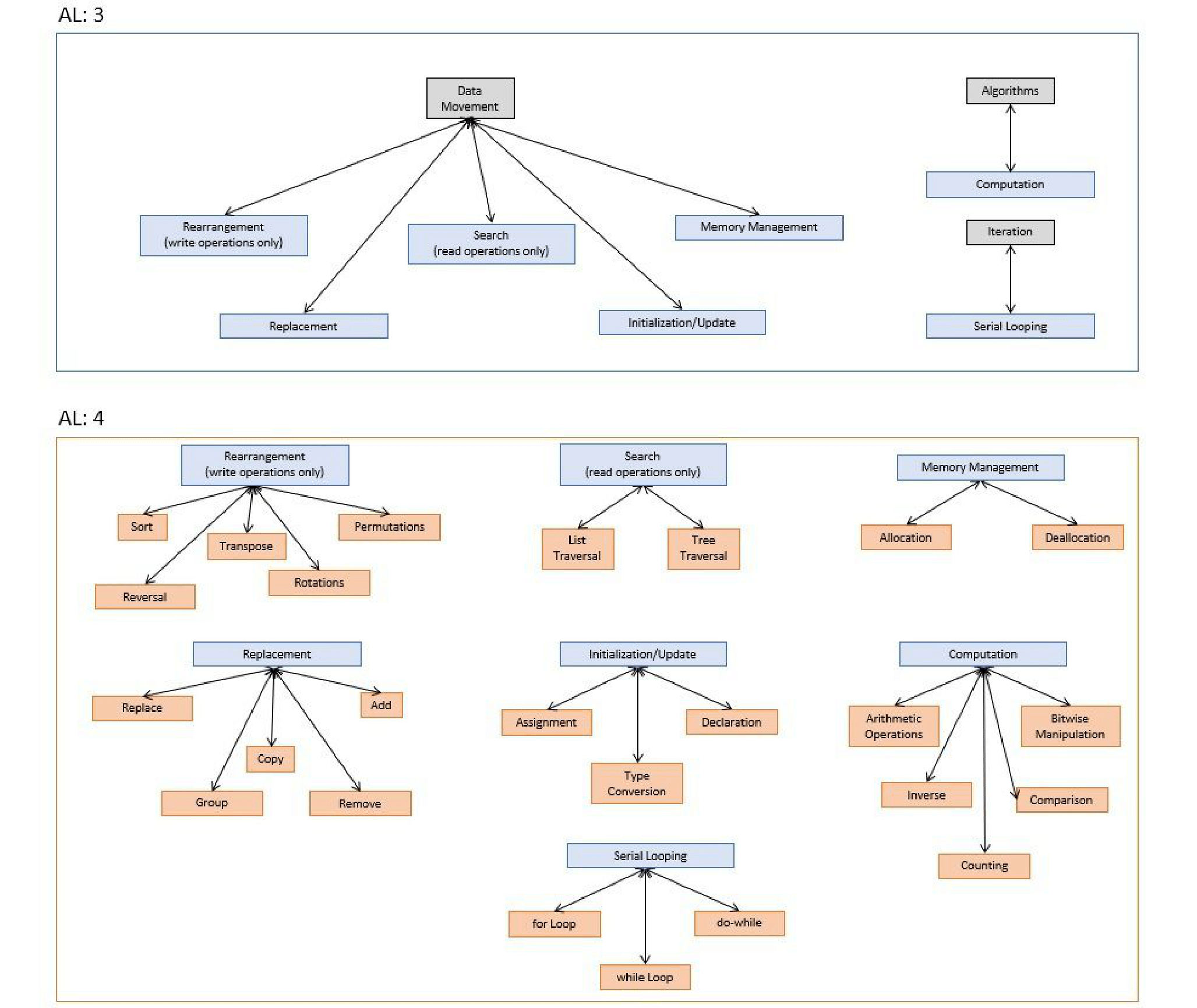}}
    % \centering
    % \includegraphics[width=16cm]{fig/fig_5_2.jpg}
\end{figure}

\clearpage

\begin{figure*}[t!]
    % \centering
    \centerline{\includegraphics[scale=0.21]{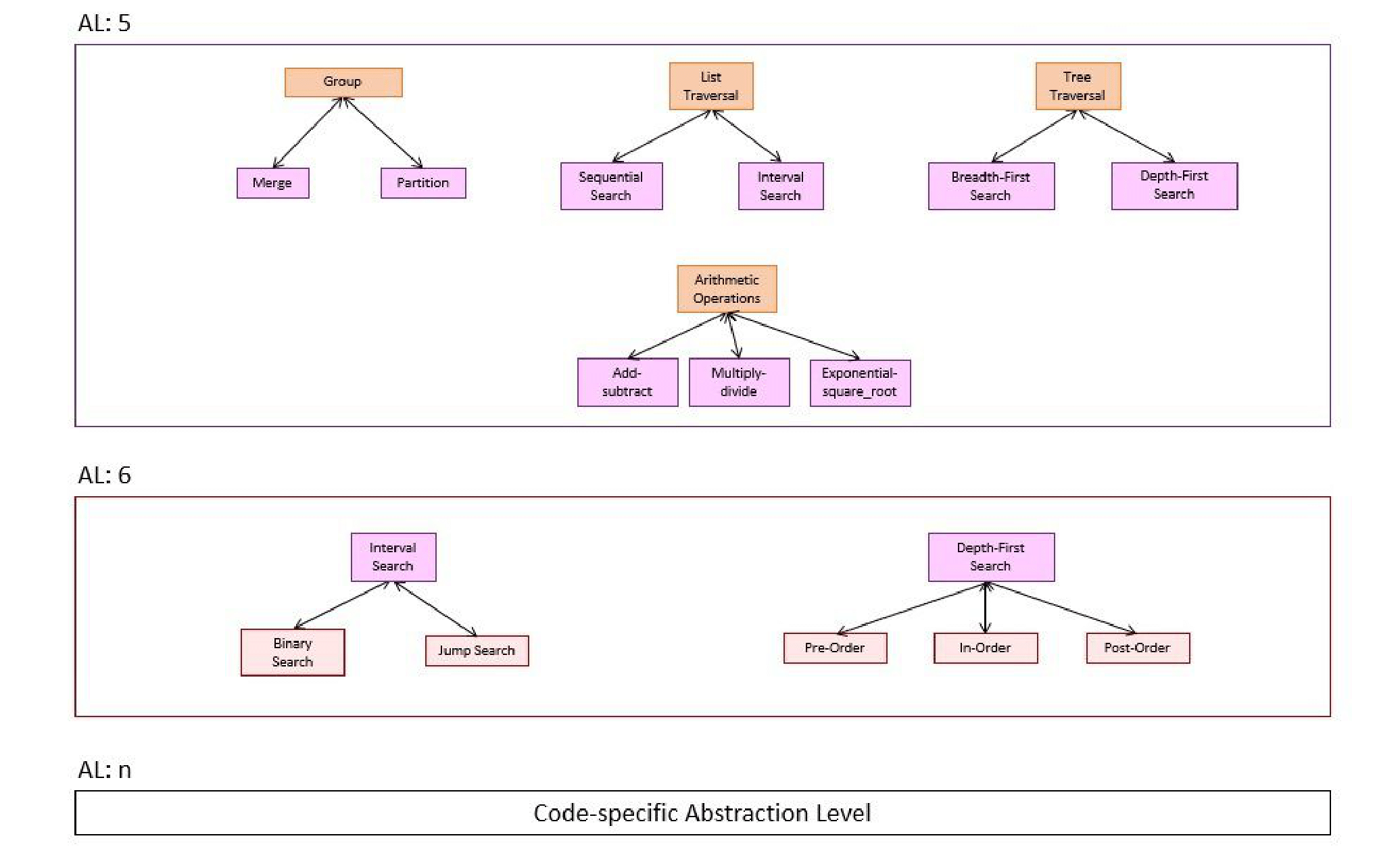}}
    \caption{\textmd{Detailed Abstraction Level (AL) construction of base \graph}}
    \label{fig:base_ASG}
\end{figure*}

\clearpage

\end{document}